# Emergence of Transfer Learning towards Specific Identification of Alzheimer's Disease – A Prospective Approach

Soumik Podder
*Department of Basic Science and Humanities*
*Institute of Engineering and Management (IEM), University of Engineering and Management Kolkata Kolkata, India*
0000-0002-8599-4635

Chandramouli Haldar
*Department of Computer Science and Engineering*
*Guru Nanak Institute of Technology*
Kolkata, India
0009-0004-9759-194X

***Abstract*— Worldwide, millions of senior citizens are suffering from Alzheimer disease abbreviated as AD, a well-versed form of dementia. AD is featured by amnesia, intellectual disability, and difficulty with consciousness. DL and ML models are undoubtedly explored to identify AD related patterns on large dimensional neuroimaging data but they need global optimization and are suffering from overfitting issue that might yield dissatisfactory result in testing data set. DL overcomes the issue by convolution of input image with kernel but any sudden change in the MRI image or human manipulation, limited pre-processing of the images can mislead CNN in achieving highly accurate detection. Transfer Learning (TL) has proved itself in AD diagnosis by utilizing pre-trained models on large data sets to guide novice model in a new neuroimaging dataset. This review provides an inclusive glimpse of TL implication in classification, identification including the conversion of AD. Keeping in view, we have assessed the strengths and limitations of TL in improvising diagnostic accuracy even with limited data. The uniqueness of the present review is the incorporation of explainable AI in TL based AD diagnosis system. Finally, it can be claimed that the review will guide the new re-searchers in the area of TL induced neurodegenerative disease detection.**

## I. INTRODUCTION

Worldwide, millions of senior citizens are suffering from Alzheimer disease abbreviated as AD, a well-versed form of dementia. Presently, no disease-modifying treatment is available. As of 2022, nonetheless around 60 million people around the globe are symptomized with AD or supplementary forms dementias. Around 60–70% of dementia cases are treated as AD [1]. AD is featured by unexpected symptoms like memory loss, sharp decline in intellectual ability etc. thereby creating a burden on patients' caretakers and society. Generally speaking, dementia spectrum is partitioned by NC, MCI, and AD. The key concept in identifying AD is to detect MCI, an intermediate state between AD and NC. MCI is symptomized by substantial amnesia, leading to hypomnesia. MCI is suspicious in terms of daily activity management but shows a sharp decline in cognitive abilities. Therefore, proper detection of MCI/AD is important, and the classification of MCI and AD is highly needed for the optimum treatment of dementia. Moreover, early detection and medication of MCI can inhibit the progression to AD. The transformation of MCI to AD engraves more than classification, with very nearly 35% of patients with MCI being vulnerable to AD within 3 to 5 years [2]. Thus, quick and accurate multidimensional detection of AD is highly sought by medical practitioners. Unfortunately, conventional medical diagnosis systems are bottlenecked by the limited knowledge of doctors and nurses. Computer-enabled diagnosis offers automatic classification and prediction of AD. There are several open-source databases for AD [3,4] among which ADNI [5], AIBL, OASIS, J-ADNI [6,7] are the well-known and commonly used database. Imaging technologies have proven to be the best method for identifying the evolution of AD. MRI (Fig. 1) as well as Positron Emission Tomography (PET)-these two imaging technologies are established techniques for identifying AD, as these two imaging technologies are associated with biomarkers such as amyloid-beta and tau proteins. Currently, MRI-related software such as Statistical Parametric Mapping (SPM) and Freesurfer are freely available and widely used by researchers to identify AD. However, Computer-Aided Software still lacks the accuracy needed to detect the conversion of pMCI from sMCI. DL and ML models have been extensively explored to identify AD-related patterns in large neuroimaging datasets. Shallow ML models like SVM and RF are employed, but they face issues with global optimization that do not fit the problem well. On the other hand, ANN is used for local optimization. In advanced ANN forms, deep ML models such as CNN and RNN are extensively used to accurately identify AD from neuroimaging data. DL models inherently include feature extraction steps within the learning model itself. Ensemble methods are also a competitive tool for detecting AD, though they face challenges such as overfitting and underfitting of data [8]. On the contrary, transfer learning (TL), a DL approach, effectively addresses the problem of insufficient labelled neuroimaging data, advancing the detection of AD accurately. By using pre-trained features, a smaller amount of data is needed, drastically cutting down computation time to train the model for a new task. TL further fine-tunes the pre-trained weights of models, enhancing performance for a given classification task. For instance, pre-trained 3D CNN models like ResNet or VGG16 can now be used to differentiate between pMCI and sMCI. These models are built for general image classification or AD classification [8]. This review provides a chronological development of ML models, starting from shallow models and progressing to deep learning models, and finally addresses recent advances in TL for diagnosing AD. TL augments the accuracy of AD diagnosis early by exploiting pre-trained models. TL encompasses the synergistic effect of speech and NLP along with analytical tools used in MRI scanned images. The main mechanism of TL is to transfer knowledge from other domains, which is efficient in producing promising results in classifying and diagnosing AD. Despite of all advantages, challenges, limitations, and knowledge gaps still remain, which are systematically addressed in this review article. Finally, the review may help researchers by providing lines of inquiry that future research can take, in advancing the solicitation of TL based AD detection.

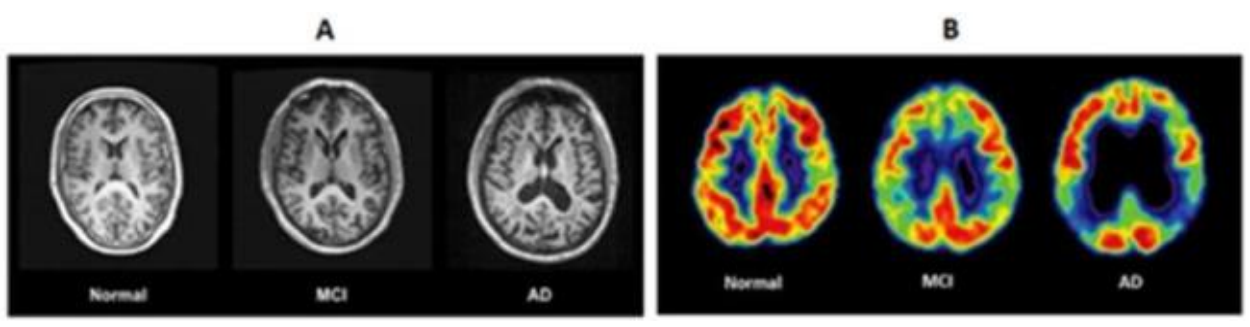

Fig. 1. A ) MRI scanned image of various stages of AD; B) PET scanned image of different AD stages such as normal, MCI, and AD.

## II. Employment of Shallow Models at AD Detection

Shallow ML models are used in AD diagnosis as they are stable, widely used, and well-established methods. Among shallow ML models, support vector machines (SVMs) are a popular model often used in AD detection due to their stability, structural risk minimization principles, and generalization performance. [8] For predicting AD from an MRI image, image modality is a very important factor. The most commonly used modality is T1-weighted images for structural MRI, but T2-weighted images are also exercised by a few researchers. This is because T1-weighted images clearly illuminate the delineation of the ventricular surface of the brain owing to atrophy. Using feature selection and extraction methods such as demographic and genetic information conjugated MRI, parcellation method for identifying gray matter changes, histogram of regions of interest (ROIs), normalized mean square by SVM, and Fischer discriminate ratio (FDR), detecting MCI and AD becomes more accurate, as evident from many researchers [9-13]. The wavelet-based feature extraction method is also a competitive method for classifying AD images from control null images, as evident from the application of discrete wavelet transform features (DWT) at SVM in classification of control null, MCI, and AD images. SVM with EEG data also performed well in classifying AD from control MRI images. Speech patterns with SVM classifiers exhibited excellent performance in identifying AD from MRI images.
[13] In SVM, different kernel functions like linear, polynomial, RBF, are also used in recognition of AD, and leave one out cross validation (LOOCV). Among ensemble methods, random forest (RF) is an established and popular model as it consists of multiple decision trees that act as individual classifiers. RF takes the most voted classification result after taking the entire result from parallel processing of multiple decision trees' classifications. The disadvantage of SVM over large-dimensional data is theoretically challenged by RF as it works fantastic in the sense of zero-dimensional reduction.

## III. Emergence of DL in AD Detection

DL comes into the stage as DL models function in the classification or prediction layer-wise, yielding higher accuracy and overcoming overfitting effects as laid down by ML models. CNN is a variety of DL that deals with input datasets spatially (input data is in the form of a matrix). The spatial arrangement of data helps in more accurate detection. In CNN, convolutional operations take place between the input data matrix and the filter (kernel) to extract useful features. CNN is widely used in organ segmentation and disease detection, especially in cancer cell identification. In general, CNN structure consists of cascaded convolutional layers, filters, stride, padding, and fully connected networks. The activation functions used in CNN are sigmoid, ReLU, Leaky-ReLU, and tanh functions. The convolution process entertains sliding the filter from top to bottom, left to right to reduce the dimension of the original input image. Numerous reports claim CNN-based architecture as a magnificent solution to detect AD as well as the conversion of MCI to AD. [8] Salehi et. al. has illuminated the diagnosis of AD and classified AD from MRI images consisting of a total of 2633 non-demented, 2480 AD and 1512 mild demented categories from the ADNI database with the help of CNN. The CNN model exhibited almost 99% accuracy in the detection of AD. They have also pointed out that shallow ML models are not suitable for large datasets as collected from the OASIS database [14]. Assy et. al. concatenated two simplified CNN architectures in a 5-way classification task to improve the accuracy from 95% to 99.13%, and the model had task-specific features and complementing characteristics [15]. Another CNN-based approach is found in Helaly et. al.'s report where 2D and 3D MRI images are used as testing datasets. The accuracy to detect multiclass AD for 2D and 3D images is 93.61% and 95.17% respectively [16]. The CNN architecture-based approach outperforms because of notable contributions of LeNet-5, HadNet, and 3D multi-scale frameworks in AD diagnosis.

## IV. Recent Development in AD Diagnosis and Its Challenges

### A. Introduction to TL

As CNN architecture performance depends on image filtering, hence any sudden changes in the MRI images of brain eventually affects the performance of CNN framework. Dementia affected patients experiences sudden changes in brain anatomy and those intricate symptoms defy CNN excellence. Even limited pre-processing of the images and human manipulation often lead to misguiding the CNN so the recognition and discrimination of AD and MCI may produce error message. CNNs with higher number of layers are subjected to immense research but no guaranteed improvement is observed. For solving such unpredicted situation and consecutive complex pattern of MRI images, transfer knowledge from another data set might be helpful as this idea has been developed from human knowledge transfer between tasks. The advantage of such knowledge transfer is faster workflow with success. This transfer of knowledge between tasks is motivated from NIPS-1995 workshop

"Learning to Learn". The theme of the workshop was then emphasized by Andrew Ng in a tutorial named NIPS 2016 in the year of 2016.[42] The tutorial predicted that TL is the forthcoming future of ML. TL is aimed in knowledge consolidation and transfer. It eliminates the bottlenecks of supervised and semi-supervised learning models by transfer knowledge between different domains or tasks. The operational principles of the above-mentioned framework rely on two parameters, namely 1. Domain and 2. Task. Domain consists of a feature space $\chi$ that comprises an n-dimensional feature vector indicating multidimensional properties of an object and marginal probability distribution function P(X) where X = {X1, X2, X3, …., Xn} ∈ $\chi$. Task represents the label space Γ consisting of labeled properties of training objects and an objective function f: $\chi \rightarrow \Gamma$. Now, correlating with the above figure, the domain represents data sets 1 and 2 as they provide n-dimensional features. (Fig. 2) The Learning System Task has two departments, such as model and the label space. The model executes itself according to its objective function. Using this approach, CNN can work well without a large data set. The time complexity in the CNN based detection process becomes drastically reduced after the induction of the TL mechanism, but one limitation in the TL based approach is that if the previously trained model is assigned a new operation, then a specific layer needs to be retrained, keeping other layers retained in their previous learning state. This situation leads to two circumstances: 1. Which layer should be retrained? 2. Which layers will be frozen?[17].

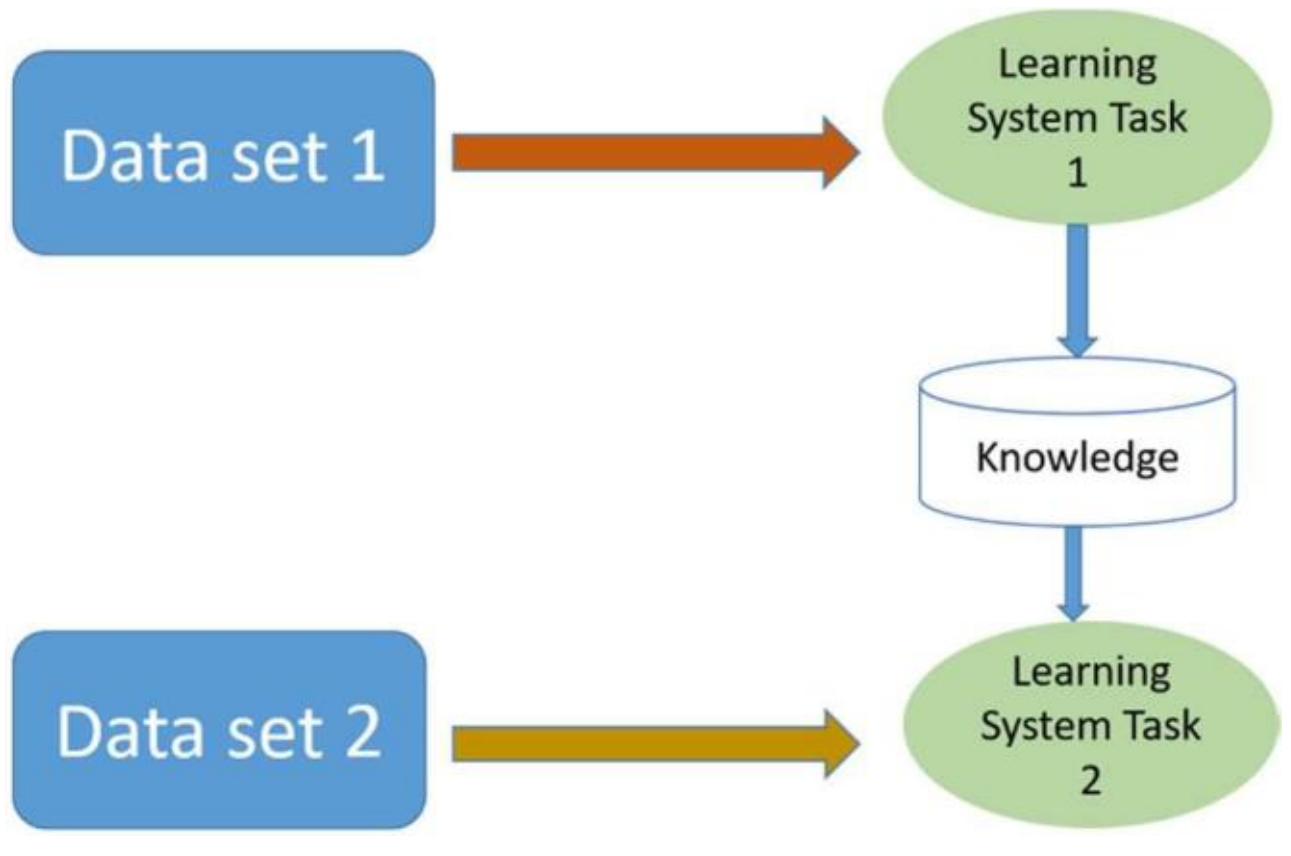

Fig. 2 Framework of TL.

### B. Implementation of TL in Early Diagnosis of AD

TL has transformed AD diagnosis by considerably increasing the accuracy of neuroimaging-based categorization.TL architecture offers less time of computation and comparatively noteworthy performance. Pre-trained convolutional neural networks (CNNs) such as VGG16, ResNet-50, deep ResNet, CaffeNet, DenseNet and AlexNet, which were previously trained on large-scale picture datasets such as ImageNet, have been repurposed for AD diagnosis. When fine-tuned on neuroimaging datasets, these networks exceptionally should be used and, conversely, for example, combining TL and MRI data resulted in an outstanding classification accuracy of 95.70% accurately separating AD patients from healthy controls. This demonstrates TL's capacity to reuse generic picture characteristics from nonmedical datasets and apply this knowledge to medical tasks with great accuracy [18]. Mehmood *et. al*. had scripted a very interesting observation against TL mediated AD detection at early fashion. The detection was carried out gray matter scans as indicative feature of AD detection. They developed layer-wise learning model perfectly realized by VGG architecture family with previously trained weights for AD cataloging. Binary classification including NC, EMCI, LMCI and AD were performed to achieve best performance as evident from accuracy results. The as obtained accuracy of detection AD from NC is 98.73% whereas EMCI patients were distinguished from LMCI patients at an accuracy level of 83.72%. The residual binary classification was performed at accuracy level of 80%.[19]

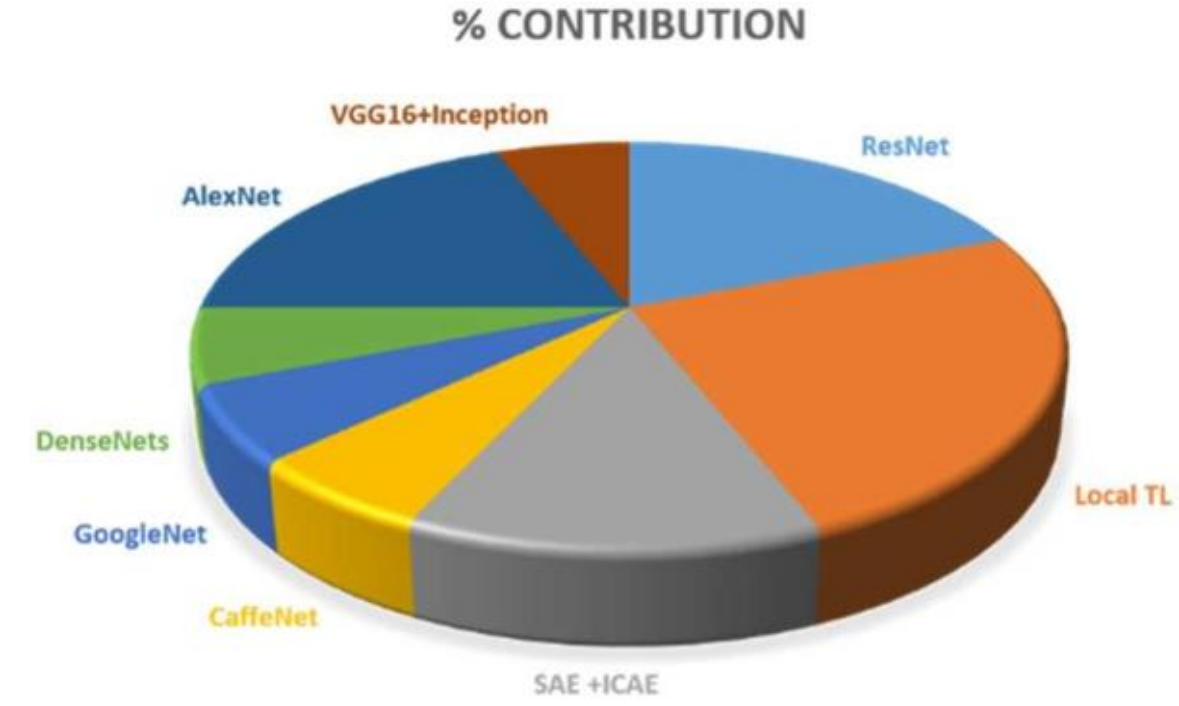

Fig.3 Distribution of different types of DL+TL architectures in AD diagnosis [19]

### C. Facilitation of Fine Tuned Pre-Trained Models for Better Precision

Studies suggest that progressively fine-tuning pre-trained CNN models across numerous epochs enhances diagnostic precision. For example, fine-tuning across 50 epochs resulted in a diagnostic accuracy of 97.84%, which was much greater than the initial phases of training. This consistent progress illustrates TL's potential for high precision in AD diagnosis provided the training procedure is well regulated and adjusted [20].

### D. TL-CNN architectures for feature extraction

CNN-based Transfer Learning architectures are proven to be operative in the analysis of AD by collecting features automatically from MRI and PET data. These designs, which have layers designed to capture complicated patterns, allow us to avoid human feature engineering, which is time-engulfing and likely to be erroneous. These models excel in detecting subtle neuron generation processes associated with Alzheimer's disease by leveraging deep features learnt from large-scale datasets. Another wing of TL induction in AD diagnosis is combating with annotation in data set. Such type of challenge creates difficulty in diagnosis of MCI at early stage and eventually the alteration of MCI to AD. However, such type of uneven fact can be conquered by Siamese 4D-Alznet, a novel CNN frame- work consisting of 5 CNN layer blocks and TL architectures such as customized AlexNet, Frozen VGG-16 as well as Frozen VGG-19. The accuracy of AD vs NC is 89% and AD vs MCI has been improved by 7%. [21] For example, models such as ResNet- 50 and Inception V3 have been demonstrated to accurately collect and transmit key characteristics for illness categorization [21].

Conjugation of different DL models with TL architectures for AD binary classification is depicted in Table 1

TABLE I. CONJUGATION OF DIFFERENT DL MODELS WITH TL ARCHITECTURES FOR AD BINARY CLASSIFICATION

| SN | Models Used | Year of Publication | AD vs NC Efficiency (%) | References |
|---|---|---|---|---|
| 1 | 2DCNN +ResNet-18 | 2019 | 97.92 | [22] |
| 2 | 3DCNN+LocalTransferLearning | 2019 | 98.20 | [23] |
| 3 | 3DCNN+ CAE+ LocalTL | 2020 | 85.24 ±3.97 | [24] |
| 4 | Deep CNN + LocalTL | 2018 | 96.00 | [25] |
| 5 | DenseNets +RNN +LocalTL | 2019 | 89.10 | [26] |
| 6 | ResNet50, ResNet18, ResNet101 | 2025 | Baseline: 99% Overall:83% | [27] |
| 7 | HEMRD TL | 2023 | 99.1 | [28] |

### E. *Combining TL and Multimodal Data*

Recent advances in TL have expanded to incorporate the utilization of multi-modal data, merging MRI, PET, and genetic information (such as APOE genotypes) to improve diagnosis accuracy.[21] Recent works are evidenced as employment of deep transfer learning models such as VGG19, Xception, Resnet50, EfficientNetB7, DenseNet201, and, hybrid EEG induced fused CT-MRI based robust principal component analysis integrated deep TL (HEMRDTL) model to early diagnosis of AD from CT-MRI images. The validation accuracy for these models is more than 90%. Pretraining of these models with Image Net dataset are carried out for extraction of the brain features from MRI images efficiently. Also, EEG data are representing functional properties of brain that are also used to detect AD in supportive fashion. [20,28]

### F. *Explainability in TL Models Ingress of Explainable AI*

A key area of recent progress has been the emphasis on making TL models more interpretable. Given that healthcare practitioners rely on clear and intelligible diagnostic tools, attempts have been made to include explainable AI approaches into TL models. This allows doctors to understand why the model makes a certain judgment, increasing trust in AI-based diagnoses. The as designed TL model cannot answer the following questions: I. “Why did you classify that item as class A, II. why the data will not be in class B?”, III. “When will the model be succeeded or failed?”, IV. “By which process I can rectify the erroneous feature selection?”, V. “Which governing feature will be needed for training the model?”, VI. “Can the prediction be trustable?” [29]. Explainable AI (XAI) can answer according to following manner: ”I know the reason for selection”, ”I know to rectify the wrong feature selection”, ”I can rely on the prediction you gave” [21,29]Trending research on XAI reveals different DL and ML models are incorporating compatible XAI technologies in a growing manner, e.g.: RF, Adaboost, XGBoost are employing Ll, Ma, Ph, Gl, methods for identification of AD from HC and MCI cases, AD vs HC, HC against EMCI vs LMCI vs AD respectively, DL architectures like 3D VGG16, 3D CNN, VGG16 use Ll, Gl, Ph, Ma for AD vs HC, HC against AD, ND against VMD vs MiD vs MoD based on sMRI, T1-Weighted, Preprocessed, Baseline MRI, T1-Weighted MRI input modalities. TL architectures such as VGG19, ResNet50, Densenet121, and InceptionV3 models are hybridized with XAI: GradCAM model to enhance the confidence of the ensemble TL model resulting in increasing adoption and steadfast outcome. The application of heatmaps and attention processes, for example, enables the depiction of key areas in neuroimaging scans, offering introspection into the model’s decision-making process flow. [21,29,30,31]

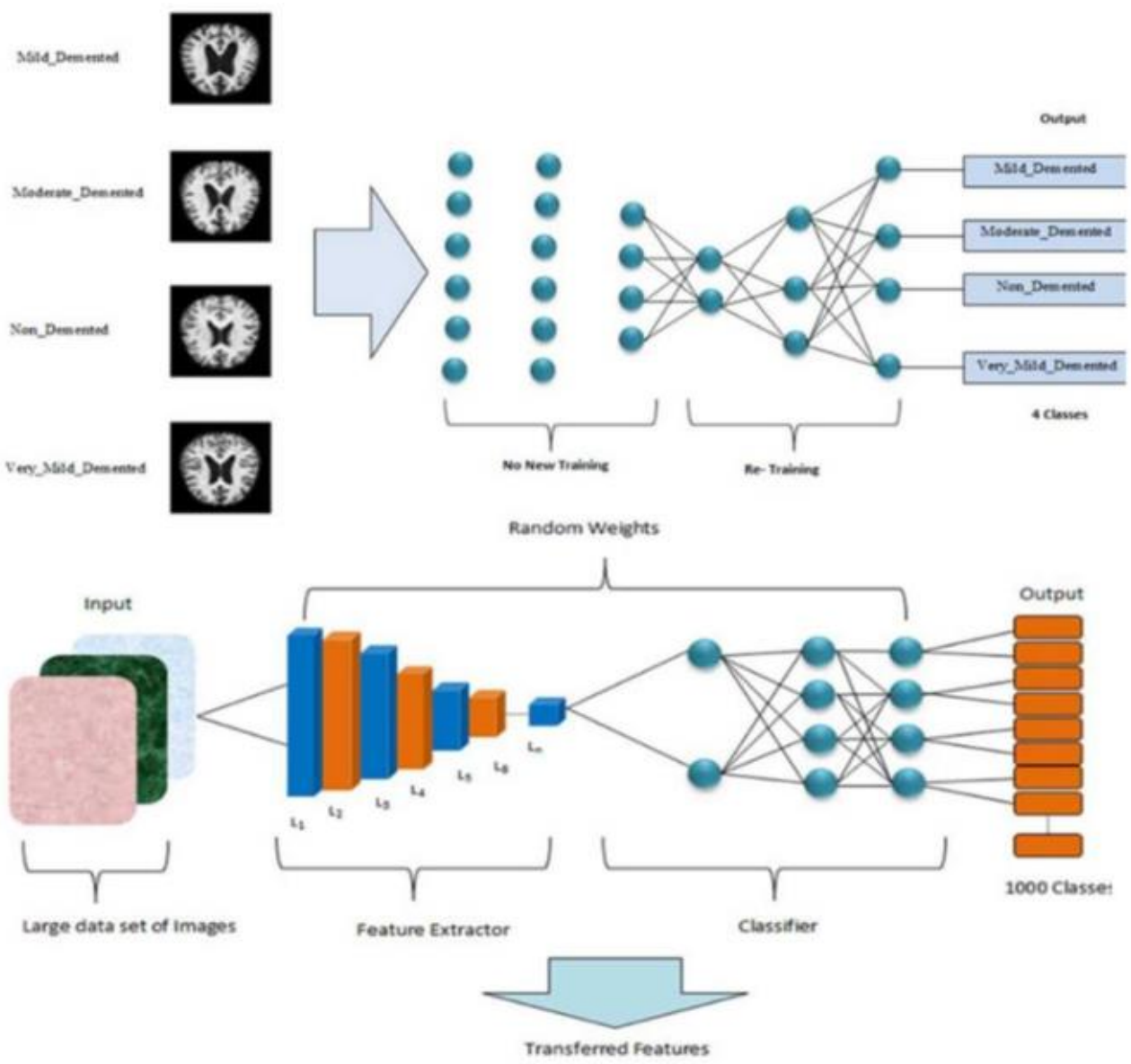

Fig. 4 Transfer Learning Framework for discrimination of different AD stages. [28]

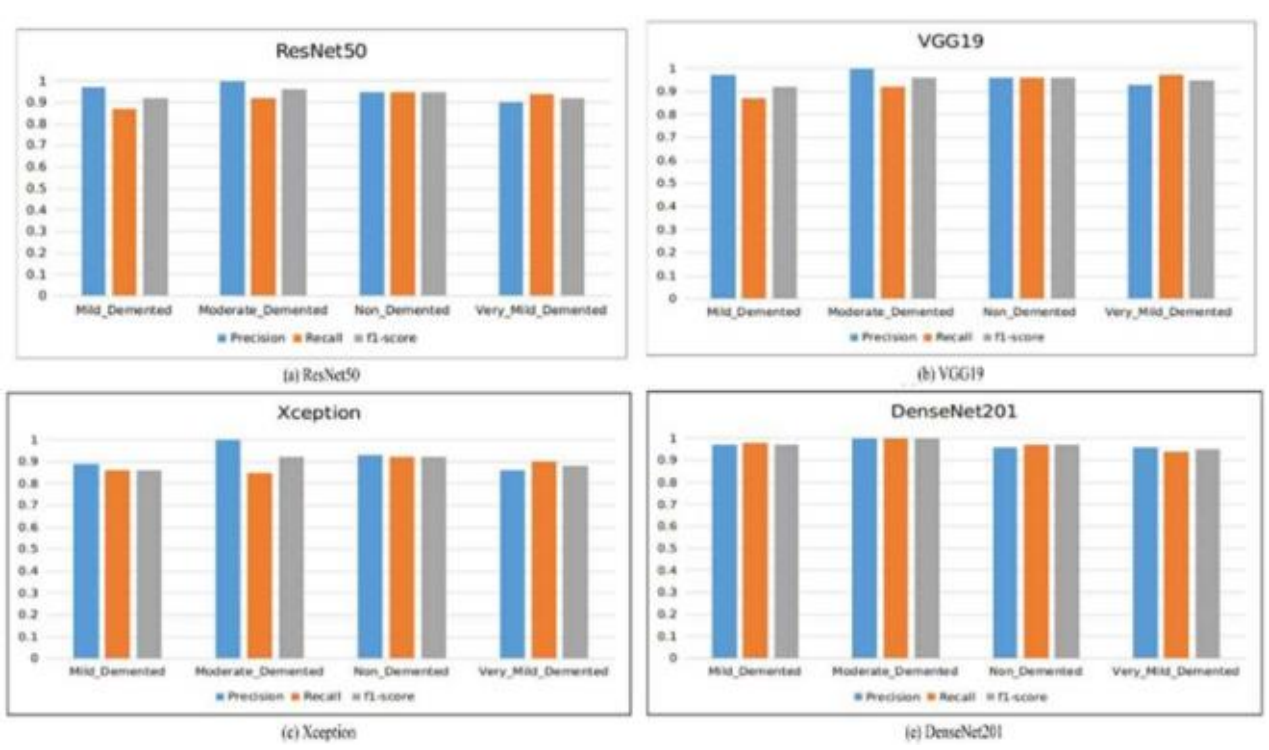

Fig. 5 Precision, Recall, F1score of ResNet50, VGG19, Xception and DenseNet201 models [28]

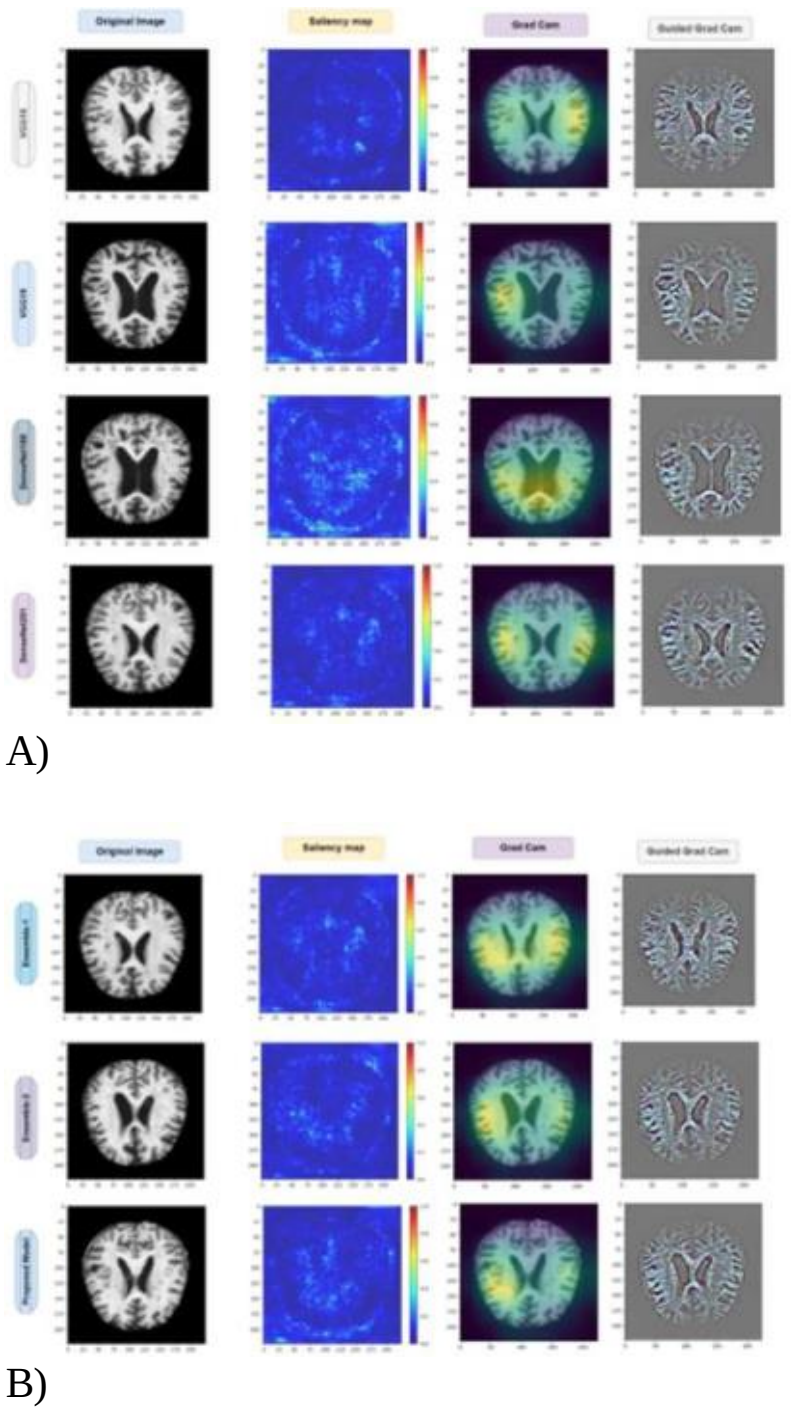

A)

B)

Fig. 6 Pictorial illustration of the saliency map as well as XAI: grad-CAM results for the A) pretrained model and B) ensemble model. [31]

## V. Challenges in Implementing TL

### A. Over fitting due to limited data set sizes

Although TL overcomes some of the challenges related to limited datasets, overfitting remains a key concern. Pre-trained models, even when fine-tuned on tiny, domain-specific datasets, are vulnerable to overfitting, especially when the training data is restricted in variety. Several research solved this issue by using data escalation techniques including flipping rotation, flipping, and zooming to artificially increase dataset size. However, overfitting remains a worry, especially when data augmentation procedures are insufficient to match real-world variability in neuroimaging data [21,32]

### B. Challenges of Multi- Class Classification

Alzheimer's disease diagnosis frequently necessitates not just binary classification (AD vs. healthy), but also multiclass classification, with the model distinguishing between many phases of AD (e.g., mild cognitive impairment, moderate AD, severe AD). Multi-class categorization is naturally more challenging, requiring models to learn finer differences across illness phases. While TL has shown promise in binary classification tasks, achieving high accuracy across several classes remains difficult and sometimes necessitates further fine-tuning and specialized techniques like as hierarchical categorization. [19,21].

### C. Handling Noisy and Low-Resolution Imaging Data

Noise and low-resolution pictures are key challenges in neuroimaging data, particular-ly MRI and PET scans, and they can degrade the effectiveness of Transfer Learning models. While TL can transfer learned features from high-quality datasets, it fails to generalize well when applied to noisy medical pictures. Noise reduction approaches and the use of advanced imaging techniques like as functional MRI (fMRI), which provides more detailed information about brain activity, have all been used to im-prove the resistance of TL models to image quality issues [29].

### D. Model Reproducibility and Generalization

The dependability of results is a significant challenge when utilizing Transfer Learning to identify Alzheimer's disease. Because neuroimaging data is inherently diverse, including variances in scanner types, imaging procedures, and patient demographics, it is impossible to predict if a TL model trained in one context would perform comparably well in another. The lack of repeatability impedes the practical application of TL models, since healthcare practitioners prefer trustworthy and consistent diagnostic tools. Future research should focus on developing techniques to improve model generalization over a wide range of datasets and imaging settings. [32]

### E. Computational Complexity and Training Time

Learning models, particularly those based on deep neural networks, are computationally intensive and need significant hardware resources for training and fine-tuning. Many institutions, particularly those with limited access to high performance computing technology, may find that implementing TL models for AD diagnosis is too costly. To address this issue, more efficient model designs and approaches like as pruning and quantization have been used, with the goal of reducing computing bur-den while retaining model accuracy [18,32]. Transfer Learning has significantly advanced the area of Alzheimer's disease diagnosis, allowing for more accuracy, precision, and the ability to handle tiny datasets. The use of CNN architectures, fine-tuning approaches, and multimodal data integration has improved the diagnostic capabilities of these models. However, issues like as overfitting, noisy imaging data, and the necessity for repeatability and generalization must be solved before Transfer Learning can be completely used in clinical settings. More study is needed to address these issues and fully realize the promise of Transfer Learning in AD diagnoses.

## VI. Conclusion

The present review provides a comprehensive glimpse of the implication of TL in automated identification and discrimination of Alzheimer's Disease (AD). Several neural networks (CNNs) such as VGG16, ResNet-50, deep ResNet, CaffeNet, DenseNet and AlexNet are highlighted in binary as well as multiple classification inspired AD detection. These models when are conjugated with local TL model immediately enhance the detection accuracy due to fine tuning of top layers. Another important observation is that, the employment of 3D CNN with local TL showed better performance than 2D CNN + TL models due to availability of spatial arrangement of pixels in MRI scan. Finally, we have assessed the strengths and limitations of TL in improvising diagnostic accuracy even with limited data. Deep TL models such as VGG19, Xception, Resnet50, EfficientNetB7, DenseNet201, and, hybrid EEG induced fused CT-MRI based robust principal component analysis integrated deep transfer learning (HEMRDTL) models are now investigated to early diagnosis of AD from CT-MRI images and EEG data. The discrimination between mild- demented, non-demented, moderate demented, very mild demented MRI images are received with more than 90% validation accuracy. Although TL provides less computation time along with handling less amount of data, still there is a scarcity in clear understanding and reasoning of this AI based AD detection. In this regard, Explainable AI (XAI) models such as GradCAM, L1, G1, P1, M1 are added to provide a transparent and strong administered AD detection system. Thus, the incorporation of explain-able AI in TL based AD diagnosis system enhance the confidence, adoption and reliable outcomes. Conclusively, it can be claimed that the review will guide the new researchers in the area of TL induced neurodegenerative disease detection.

## REFERENCES

[1] World Health Organization (WHO), "World Health Organization," accessed: Oct. 09, 2024. [Online]. Available: https://www.who.int.

[2] D. Agarwal, G. Marques, I. de la Torre-D´ıez, M. A. Franco Martin, B. Garc´ıa Zapira´ın, and F. Mart´ın Rodr´ıguez, "Transfer learning for Alzheimer's disease through neuroimaging biomarkers: A systematic review," MDPI, vol. 01, pp.7259, October 2021.

[3] C. R. Jack. J, Barnes. M. A. Bernstein, B. J. Borowski, J. Brewer,S. Clegg, A. M. Dale, O. Carmichael, C. Ching, C. DeCarli, et al., "Magnetic resonance imaging in Alzheimer's disease neuroimaging initiative," Alzheimer's & Dementia, vol. 11, pp. 740–756, July 2015.

[4] M. W. Weiner, D. P. Veitch, P. S. Aisen, L. A. Beckett, N. J. Cairns, R. C. Green, D. Harvey, C. R. Jack Jr, W. Jagust, J. C. Morris, et al., "Recent publications from the Alzheimer's Disease Neuroimaging Initiative: Reviewing progress toward improved AD clinical trials," Alzheimer's & Dementia, vol. 13, no. 4, pp. e1–e85, March 2017.

[5] D. P. Veitch, M. W. Weiner, P. S. Aisen, L. A. Beckett, N. J. Cairns, R. C. Green, D. Harvey, C. R. Jack Jr, W. Jagust, J. C. Morris, "Understanding disease progression and improving Alzheimer's disease clinical trials: Recent highlights from the Alzheimer's Disease Neuroimaging Initiative," Alzheimer's & Dementia, vol. 15, pp. 106-152, January 2019.

[6] M. Fujishima, A. Kawaguchi, N. Maikusa, R. Kuwano, T. Iwatsubo, and H. Matsuda, "Sample size estimation for Alzheimer's disease trials from Japanese ADNI serial magnetic resonance imaging," *Journal of Alzheimer's Disease*, vol. 56, pp. 75–88, November 2016.

[7] T. Iwatsubo, "Japanese Alzheimer's Disease Neuroimaging Initiative: present status and future," *Alzheimer's & Dementia*, vol. 6, pp. 297–299, May 2010.

[8] Z. Zhao, J. H. Chuah, K. W. Lai, C. O. Chow, M. Gochoo, S. Dhanalakshmi, N. Wang, W. Bao, and X. Wu, "Conventional machine learning and deep learning in Alzheimer's disease diagnosis using neuroimaging: A review," *Frontiers in Computational Neuroscience*, vol. 17, February 2023, Art.1038636.

[9] B. Magnin, L. Mesrob, S. Kinkingne´hun, M. Pe´le´grini-Issac, O. Colliot, M. Sarazin, B. Dubois, S. Lehe´ricy, and H. Benali, "Support vector machine-based classification of Alzheimer's disease from whole-brain anatomical MRI," Neuroradiology, vol. 51, pp. 73–83, November 2008.

[10] R. Chaves, J. Ram´ırez, J. M. Go´rriz, M. Lo´pez, D. Salas- Gonzalez, Alvarez, and F. Segovia, "SVM-based computer-aided diagnosis of the Alzheimer's disease using t-test NMSE feature selection with feature correlation weighting," Neuroscience Letters, vol. 461, pp. 293–297, September 2009.

[11] C. Plant, S. J. Teipel, A. Oswald, C. Bo¨hm, T. Meindl, J. Mourao- Miranda, A. W. Bokde, H. Hampel, and M. Ewers, "Automated detection of brain atrophy patterns based on MRI for the prediction of Alzheimer's disease," NeuroImage, vol. 50, pp. 162–174, March 2010.

[12] J. Ram´ırez, J. M. Go´rriz, M. Lo´pez, D. Salas-Gonzalez, I. Alvarez, F. Segovia, and C. G. Puntonet, "Early detection of the Alzheimer disease combining feature selection and kernel machines," in International Conference on Neural Information Processing, Springer, pp. 410–417, 2008.

[13] M. Tanveer, B. Richhariya, R.U. Khan, A.H. Rashid, P. Khanna, M. Prasad, and C.T. Lin, "Machine Learning Techniques for the Diagnosis of Alzheimer's Disease: A Review," *ACM Transactions on Multimedia Computing, Communications and Applications*, vol. 16, pp. 35, October 2020.

[14] A. W. Salehi, P. Baglat, B. B. Sharma, G. Gupta, and A. Upadhya, "A CNN Model: Earlier Diagnosis and Classification of Alzheimer Disease using MRI," in *2020 International Conference on Smart Electronicsand Communication (ICOSEC)*, Trichy, India, pp. 156-161, October 2020.

[15] A. M. El-Assy, H. M. Amer, H. M. Ibrahim, et al., "A novel CNN architecture for accurate early detection and classification of Alzheimer'sdisease using MRI data," *Scientific Reports*, vol. 14, pp. 3463, February 2024.

[16] H. A. Helaly, M. Badawy, and A. Y. Haikal, "Deep Learning Approach for Early Detection of Alzheimer's Disease," *Cognitive Computation*, vol. 14, pp. 1711–1727, November 2021.

[17] R. Caruana, D. L. Silver, J. Baxter, T. M. Mitchell, L. Y. Pratt, and S. Thrun, "Learning to learn: knowledge consolidation and transfer in inductive systems," 1995.

[18] N. Raza, A. Naseer, M. Tamoor, and K. Zafar, "Alzheimer Disease Classification through Transfer Learning Approach," Diagnostics, vol. 13, pp. 801, February 2023.

[19] A. Mehmood et al., "A Transfer Learning Approach for Early Diagnosisof Alzheimer's Disease on MRI Images," Neuroscience, vol. 460, pp. 43–52, January 2021.

[20] M. Leela, K. Helenprabha, and L. Sharmila, "Prediction and classification of Alzheimer Disease categories using Integrated Deep Transfer Learning Approach," Measurement: Sensors, vol. 27, pp. 100749, June 2023.

[21] A. Mehmood, F. Shahid, R. Khan, M. M. Ibrahim, Z. Zheng, "Utilizing Siamese 4D-AlzNet and Transfer Learning to Identify Stages of Alzheimer's Disease, Neuroscience, vol. 545, pp.69-85, May2024.

[22] F. Ramzan, M.U.G. Khan, A.Rehmat, S. Iqbal, T.Saba, A.Rehman, Z. Mehmood, "A Deep Learning Approach for Automated Diagnosis and Multi-Class Classification of Alzheimer's Disease Stages Using Resting- State fMRI and Residual Neural Networks". J. Med. Syst. vol. 44, pp.37, December 2019.

[23] K. Oh, YC. Chung, K.W. Kim, et al. " Classification and Visualization of Alzheimer's Disease using Volumetric Convolutional Neural Network and Transfer Learning". Sci Rep vol.9, pp. 18150, December 2019.

[24] J.Wen, E.Thibeau-Sutre, M.Diaz-Melo, J.Samper-Gonza´lez, A.Routier, S. Bottani, D. Dormont, S. Durrleman, N. Burgos, O. Colliot, "Convolutional neural networks for classification of Alzheimer's disease: Overview and reproducible evaluation". Med. Image. Anal. vol.63, pp.101694, July 2020.

[25] H.Choi, K.H. Jin, "Predicting cognitive decline with deep learning of brain metabolism and amyloid imaging". Behav. Brain Res. vol. 344, pp.103–109, May 2018.

[26] F.Li,; Liu, M. A hybrid Convolutional and Recurrent Neural Network for Hippocampus Analysis in Alzheimer's Disease. J. Neurosci. Methods vol.323, pp. 108–118, 2019.

[27] R.Turrisi, S. Pati, G.Pioggia, G. Tartarisco, "Adapting to evolving MRI data: A transfer learning approach for Alzheimer's disease prediction",NeuroImage,vol.307,pp.121016, January 2025.

[28] P.S. Sisodia, G.K.Ameta, Y. Kumar, et al. "A Review of Deep Transfer Learning Approaches for Class-Wise Prediction of Alzheimer's Disease Using MRI Images". Arch Computat Methods Eng vol. 30, 2409–2429, January 2023.

[29] G. Yang, Q. Ye, J. Xia "Unbox the black-box for Alzheimer's disease classification: Explainable deep learning model. Biomedical Signal Processing and Control", vol. 87,pp.104827, June 2024.

[30] F.García-Gutiérrez, L. Hernández-Lorenzo, M. Nieves C.-Martín, J. A. Matias-Guiu, J. L. Ayala, "Predicting changes in brain metabolism and progression from mild cognitive impairment to dementia using multitask Deep Learning models and explainable AI",NeuroImage,vol. 297,pp. 120695,June 2024.

[31] T. Mahmud, K. Barua, S.U. Habiba, N. Sharmen, M.S. Hossain, K. Andersson, "An Explainable AI Paradigm for Alzheimer's Diagnosis Using DeepTransfer Learning". Diagnostics,vol.14, pp. 345, February 2024.

[32] M. Heenaye-Mamode Khan, P. Reesaul, M. M. Auzine, and A. Taylor, "Detection of Alzheimer's disease using pre-trained deep learning mod- els through transfer learning: a review," Artif. Intell. Rev, vol. 57, pp. 275,October 2024, doi: 10.1007/s10462-024-10914-z.